\documentclass[conference]{IEEEtran}
\IEEEoverridecommandlockouts
\usepackage{cite}
\usepackage{amsmath,amssymb,amsfonts}
\usepackage{graphicx}
\usepackage{textcomp}
\usepackage{xcolor}
\usepackage{algorithm}
\usepackage{algpseudocode}
\usepackage{multirow}
\usepackage{tabularx,booktabs}
\usepackage{pgfplots}
\usepackage{tikz}

\pgfplotsset{compat=1.18}

\def\BibTeX{{\rm B\kern-.05em{\sc i\kern-.025em b}\kern-.08em
    T\kern-.1667em\lower.7ex\hbox{E}\kern-.125emX}}
\begin{document}

\title{Mitigating System Bias in Resource Constrained Asynchronous Federated Learning Systems}

\author{\IEEEauthorblockN{
Jikun Gao\IEEEauthorrefmark{1},
Ioannis Mavromatis\IEEEauthorrefmark{2},
Peizheng Li\IEEEauthorrefmark{1},
Pietro Carnelli\IEEEauthorrefmark{1},
Aftab Khan\IEEEauthorrefmark{1}
}\\ 
\vspace{-3.00mm}
\IEEEauthorblockA{
\IEEEauthorrefmark{1}Bristol Research and Innovation Laboratory, Toshiba Europe Ltd., U.K.\\
\IEEEauthorrefmark{2}Digital Catapult, London, U.K.\\
Email: {\{Peizheng.Li, Pietro.Carnelli, Aftab.Khan\}@toshiba-bril.com}, Ioannis.Mavromatis@digicatapult.org.uk
}}


\maketitle
\hyphenation{Distribution Heterogeneity}

\begin{abstract}
Federated learning (FL) systems face performance challenges in dealing with heterogeneous devices and non-identically distributed data across clients. We propose a dynamic global model aggregation method within Asynchronous Federated Learning (AFL) deployments to address these issues. Our aggregation method scores and adjusts the weighting of client model updates based on their upload frequency to accommodate differences in device capabilities. Additionally, we also immediately provide an updated global model to clients after they upload their local models to reduce idle time and improve training efficiency. We evaluate our approach within an AFL deployment consisting of 10 simulated clients with heterogeneous compute constraints and non-IID data. The simulation results, using the FashionMNIST dataset, demonstrate over 10\% and 19\% improvement in global model accuracy compared to state-of-the-art methods PAPAYA and FedAsync, respectively. Our dynamic aggregation method allows reliable global model training despite limiting client resources and statistical data heterogeneity. This improves robustness and scalability for real-world FL deployments.
\end{abstract}


\section{Introduction}
Federated Learning (FL) is a collaborative Machine Learning (ML) method, which aims to meet the significant challenges of (user) data privacy and the large number of devices. FL was first proposed by Google/Alphabet in 2016~\cite{mcmahan2017communication}. In traditional ML paradigms, datasets are centralised in a single location for processing and learning~\cite{naeem2020sdn}. Whereas, with an FL framework, data is distributed on different devices and no longer needs to be collected centrally. Each participating device (also known as a \textit{"client"}) will train the model independently using their local data. Once a training round is complete (usually involving several learning iterations), the client sends the gradient or parameter update of the model to a central aggregation node, known as the \textit{"parameter server"} (PS). The task of the PS is to aggregate the updates sent by all the involved clients, form a new global model, and broadcast the new global model back to each client~\cite{ferraguig2021survey}. FL framework ensures that user data never leaves the original device, and only model parameters are shared centrally, thus protecting user privacy to a certain extent~\cite{konevcny2016federated}.

\begin{figure}[!t]
  \centering
  \includegraphics[width=0.5\textwidth]
  {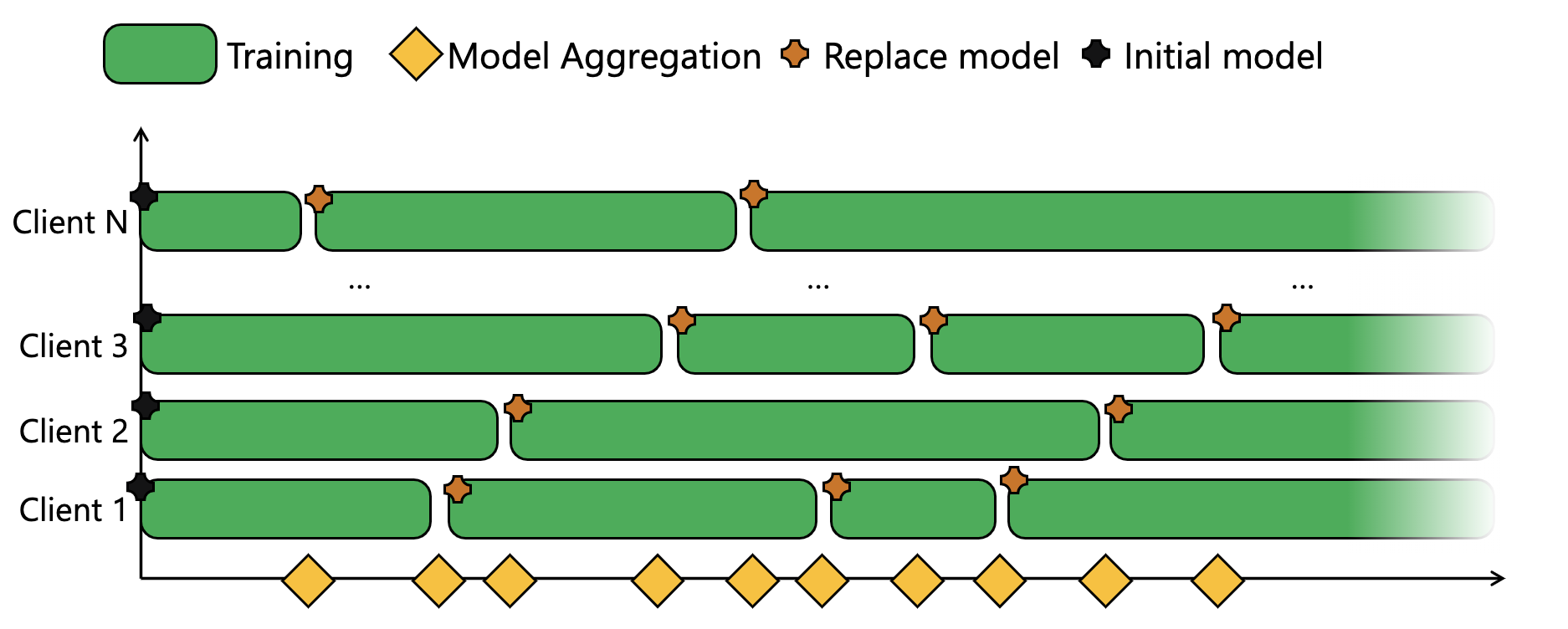}\\
  \caption{A typical AFL system overview. Starting from the left, an initialised model is sent to all clients. Clients commence training (green), and once completed, they share their model with the parameter server. Once a new client model is uploaded, it is immediately aggregated into the global model (yellow diamonds) and sent back to the client. 
  }
  \label{fig:AFL}
\end{figure}

\begin{figure*}[t]
  \centering
  \includegraphics[width=0.975\textwidth]{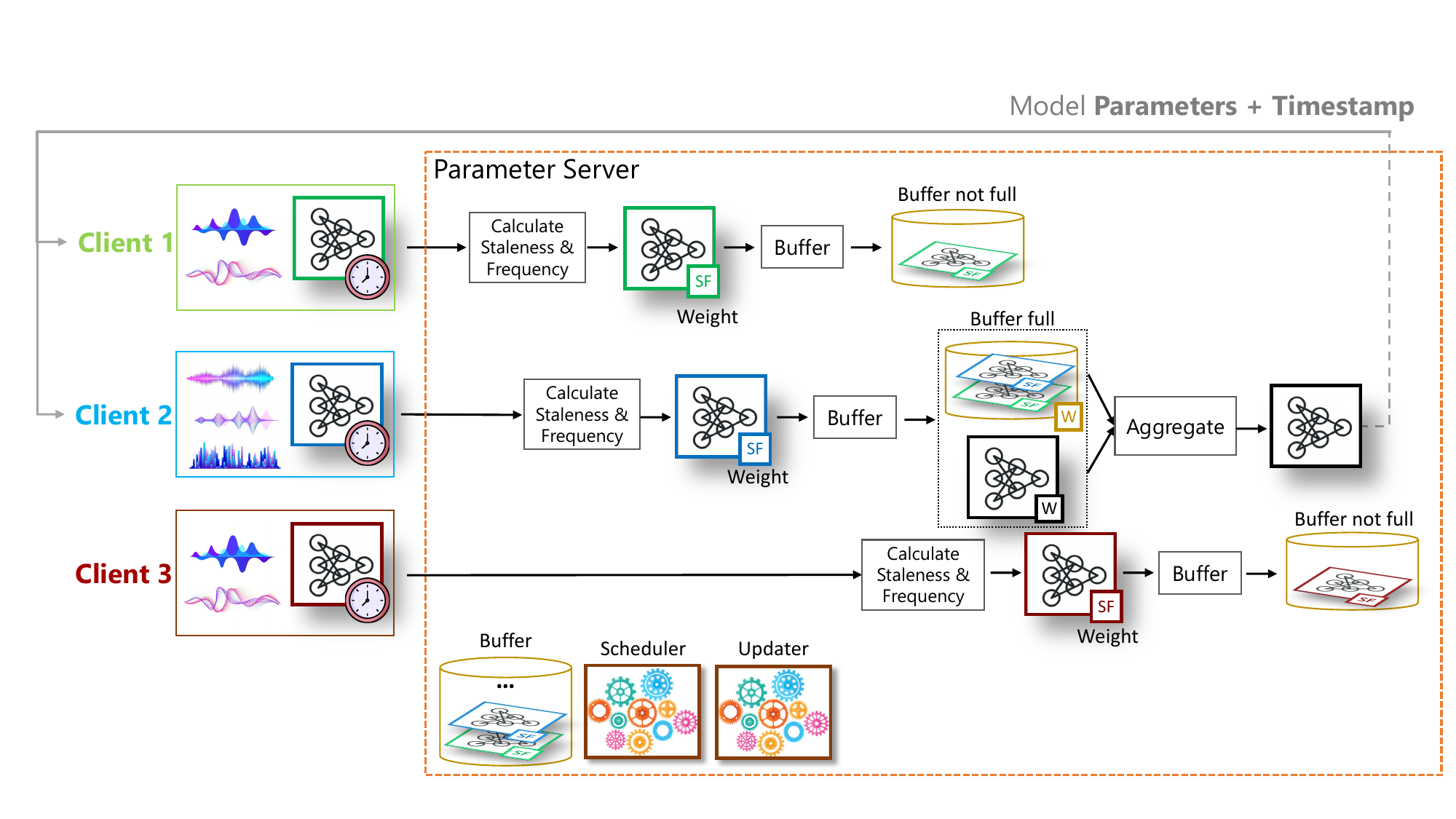}\\
  \caption{Our asynchronous FL system overview diagram. Clients train using local datasets before sharing model parameters with the Parameter Server. In turn, client models are scored/weighted according to the frequency of updates before being aggregated into a new global model for sharing with client devices.}
  \label{fig:sys-diagram-bril-async-fl}
\end{figure*}

In practical FL implementations, participating client devices exhibit heterogeneity in hardware capabilities, including communication link quality, bandwidth availability, local memory capacity, processing speed, and power constraints. In addition, given the nature of distributed devices and sensors in an FL network, the data held by participating FL devices typically follows a non-Independent and Identically Distribution (non-IID) ~\cite{li2019asynchronous}. Often these differences in hardware performance and training data distribution lead to a critical problem in real-world FL applications: how can we process the time difference in completing a round of FL training between different clients~\cite{diao2020heterofl}. In the synchronised version of FL~\cite{xu2021asynchronous}, more efficient (in terms of computational capacity) devices are often forced to wait after completing their local training tasks. This can be mainly attributed to the FL devices with inferior hardware capabilities or those with larger local datasets requiring substantial computing resources to complete client-side training procedures. Only after these devices have successfully shared their locally trained models with the parameter server can the more efficient devices proceed with the subsequent round of training~\cite{li2020federated}. This means that the efficiency of the entire system can be negatively impacted by the worst-performing client, resulting in a large number of computing resources being wasted/idling unnecessarily~\cite{li2014communication}. The distribution of non-IID data can also impede global model convergence during collaborative training as each client model drifts towards disparate local optima~\cite{zhao2018federated}. Existing mitigation techniques such as aggregation after receiving a pre-set minimum number of local models~\cite{beutel2020flower} carry inherent limitations; slower-updating edge devices face exclusion from contributions to the global model aggregation regardless of representativeness.

In order to solve the problems of delay and resource utilisation in such traditional FL settings, Asynchronous Federated Learning (AFL) was proposed (illustrated in Fig.~\ref{fig:AFL}). In this, the parameter server no longer waits for all devices or edge nodes to complete their respective training~\cite{xie2019asynchronous} prior to broadcasting a global model. Once the device completes training of its local model, it is shared with the parameter server. The server then immediately starts aggregating the new local model with the existing global model and sends the updated global model back to the device so that the device can continue the next round of training. This aggregation process also usually involves combining models from different devices. In most cases, these models are given equal weighting (when averaged) to generate a new global model~\cite{chen2021fedsa}. The advantage of this method lies in the high tolerance towards devices with different hardware capabilities and data distribution. Especially in the case of significant differences in device resources, the AFL strategy ensures that devices with superior computing performance will not waste time waiting, thus avoiding a large number of computing resources in the idle state~\cite{chen2021towards}. Since, each FL device can train a model and upload it independently, the communication or waiting with other devices is significantly reduced. This method has significant advantages in communication efficiency, especially in those scenarios with high network delay or unstable network connection~\cite{chen2019efficient}.

While AFL addresses challenges related to the operational functionality of FL deployments, the global model remains susceptible to bias from hardware heterogeneity and non-IID data distributions. Consequently, devices updating the global model more frequently may dominate aggregation, creating a skew that disadvantages resource-constrained edge devices with infrequent updates. Additionally, if not appropriately weighted for recency, stale model updates can disrupt representation. Similarly, non-IID data partitioning and class imbalance can lead to the training of disparate edge models. Naively aggregating these models may hinder convergence, resulting in suboptimal global model performance. Therefore, intelligent coordination in AFL aggregation is essential to ensure robustness against device and data heterogeneity, preventing compromised global model accuracy.

This paper introduces an enhanced AFL model aggregation method that evaluates and adjusts global model aggregation based on client upload frequency. Furthermore, it actively reduces client idling time between FL aggregation rounds by providing clients with the latest global model for training. During the FL client upload phase, a buffer layer is implemented, incorporating an aggregation scaling factor to quantify communication frequency with the server. It calculates the difference between the client model being uploaded and the global model. Dynamic parameters within the layer are configured to optimise the induced model staleness bias. Consequently, our method can train a reliable global model even when faced with limitations in the computing power of an FL client or the uneven distribution of local client training data, addressing both data volume and class imbalance.


In the subsequent sections, the paper is structured as follows: Section \ref{sec:related} covers the background to our method detailed in Section \ref{sec:methods}. We discuss our experimental setup in Section \ref{sec:experiments}. Section \ref{sec:results} shows our results which are further discussed and concluded in Section \ref{sec:conclusion}.

\section{Related Work}\label{sec:related}

In order to solve the potential bias in AFL environments, the FedAsync algorithm, proposed by Xie \textit{et al.}~\cite{xie2019asynchronous}, introduces a dynamic hybrid hyperparameter based on an outdated model. For example, when the current global model has been updated for a few rounds of aggregation and a new client has completed the first delayed local update, outdated or non-obsolete models will be queued in chronological order. The PS controls the proportion of local models when aggregating with the global model through obsolescence functions. Consequently, local models with a large degree of `staleness' will have a small weighting in the global model update. This algorithm directly reduces the impact of the stale/laggard model on the accuracy of the global model. However, these models may still contain some valuable information that could help to improve the accuracy of the global model.

Huba \textit{et al.}'s Papaya algorithm~\cite{huba2022papaya} implements the FedBuffer algorithm proposed by Nguyen \textit{et al.}~\cite{nguyen2022federated}. In this, after a client completes its local training, the updated local model is uploaded to the buffer layer memory first. When the buffer layer reaches the aggregation target (i.e., a pre-set number of client models), the global model is aggregated and updated to the clients. This can effectively reduce the gap between the laggard and the faster clients, but the computing resources of some powerful devices are still not optimally utilised because these clients still need to wait for the weaker devices in the buffer layer to complete the training and enter the idle state.

\begin{figure}[!t]
  \centering
  \includegraphics[width=0.5\textwidth]
  {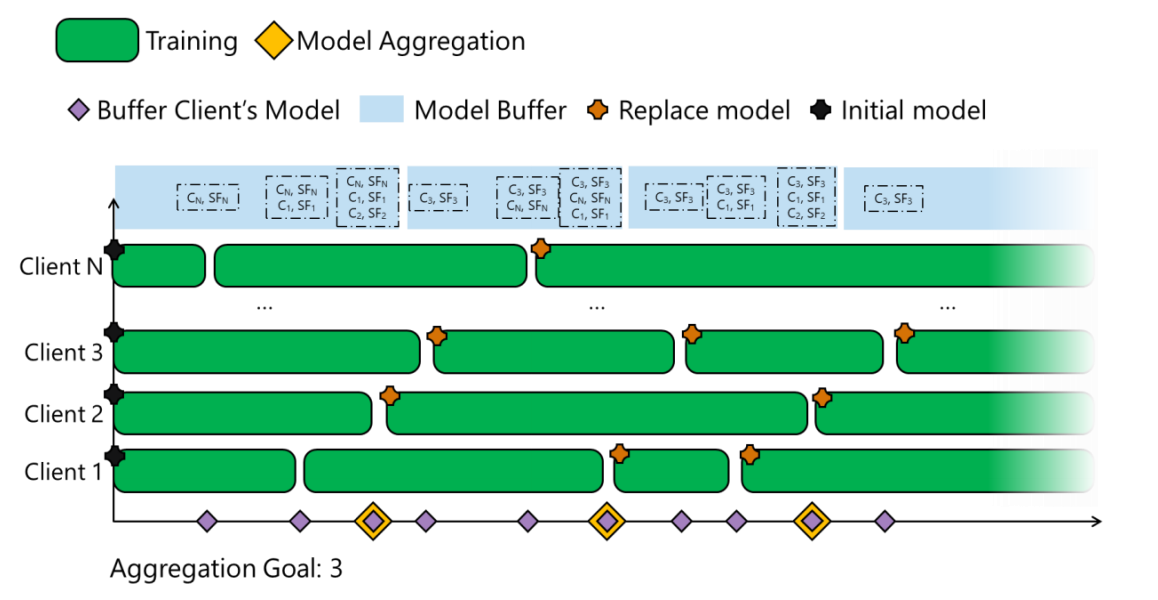}\\
      \caption{Diagram of our proposed method with an aggregation goal of 3 client models. Clients immediately receive the latest aggregation/global model once they upload their local model to the PS. However, we store and score the incoming client models in the PS buffer layer to reduce biases when aggregated. If the new global model is sufficiently different (once the buffer is full) it is then re-shared with all the clients. 
      }
  \label{fig:BAFL}
\end{figure}

At present, to the best of our knowledge, there is no method that can reduce the impact of aggregating stale models with the global model accuracy whilst also optimising for the computing resources of FL clients. In this paper, we propose a method to mitigate the impact of client hardware and data heterogeneities on the global model.

\section{Methodology}\label{sec:methods}


\begin{table}[ht]
\label{tab:notation}
\centering
\caption{Table of notation.}
\begin{tabular}{lp{6.7cm}}
\toprule
\textbf{Notation} & \textbf{Description} \\ 
\midrule
$N$ & Total number of clients \\
$[n]$ & Set of integers $\{1, \ldots, n\}$ , to represent each client\\
$T$ & Total number of global communication rounds \\
$t$ & Timestamp of the current number of global communication rounds \\
$\tau$ & Timestamp of the current local model based on which round of global model \\
$x_t$ & Global model in the $t$h global communication round on server \\
$D_n$ & Dataset on the $n$th device \\
$\alpha$ & Global model scale factor \\
$\beta_n$ & Intra-group model scale factor of $n$th client \\
$B$ & Total number of intra-group clients, equal to the buffer aggregation target \\
$[b]$ & Set of integers $\{1, \ldots, b\}$ , to represent each intragroup client\\
$t - \tau$ & Directly obtained staleness \\
$s$ & Function of staleness  \\
$f$ & Model upload frequency \\
\bottomrule
\end{tabular}
\end{table}

Our proposed framework for mitigating system bias, as introduced earlier, follows the asynchronous FL setup. A system overview diagram is depicted in Fig.~\ref{fig:sys-diagram-bril-async-fl} and Fig.~\ref{fig:BAFL} illustrates some example training rounds of our method. When the buffer layer at the PS receives a new local model from the participating FL client devices, and the buffer layer does not meet the aggregation target (i.e., the minimum number of client models in the buffer layer, prior to triggering a global model aggregation task), it immediately shares the current global model with the client device to continue the training task. Upon reaching the aggregation target, the buffer layer promptly aggregates the new global model, replacing the previous global model. This ensures the maintenance of a unique global model throughout the FL training process

\begin{algorithm}
\caption{Proposed method - Dynamic weight buffered}
\label{algorithm2}
\begin{algorithmic}[1]
\Procedure{Server}{$\alpha \in (0, 1)$}
    \State Initialise $x_0,x_{new},\alpha,b \leftarrow 0,B,t \leftarrow 0,T$
    \State Run \textsc{Scheduler()} thread and \textsc{Updater()} thread asynchronously in parallel
\EndProcedure
\Procedure{Scheduler}{}
    \State Periodically trigger training tasks on all clients and send the global model with a timestamp 
\EndProcedure

\Procedure{Updater}{}
    \For{$t \in [T]$}
        \State Receive the pair $(x_{\text{new}}, \tau)$ from any client,buffered
        \State $x_b \leftarrow$ buffered $x_{\text{new}}$
        \State $\beta_b \leftarrow$ Calculate the intragroup scale factor corresponding to each client staleness and upload frequency
        \State $x_{\text{new}}(t) \leftarrow x_{\text{new}}(t) + \beta_b * x_b$
        \State $b \leftarrow b + 1$
        \If{$b == B$}
            \State $x_t \leftarrow (1 - \alpha) * x_{t-1} + \alpha * x_{\text{new(t)}}$
            \State $b \leftarrow 0$
            \State $x_{\text{new}}(t) \leftarrow 0$
        \Else
            \State $x_t \leftarrow x_{t-1}$
        \EndIf
        \State Send $x_t$ to relative client
        \State $t \leftarrow t + 1$
    \EndFor
\EndProcedure

\Procedure{Client}{}
    \For{$n \in [N]$ in parallel}
        \If{triggered by the scheduler}
            \State Receive the pair of the global model and its timestamp $(x_t, t)$ from the server
            \State $\tau \leftarrow t, x_n \leftarrow x_t$
            \State Train based on the local dataset and upload after training is completed
        \EndIf
    \EndFor
\EndProcedure

\end{algorithmic}
\end{algorithm}

When any client device finishes uploading its locally trained model to the buffer layer, it will no longer enter the idle state but immediately obtain the latest global model for continued local training (thus avoiding computing resources idling/waiting). However, in this case, client devices with more compute resources may upload more than once in the same buffer layer (i.e., within a short time frame). Consequently, we introduced a scaling factor $\beta \in [0, 1]$ in the buffer layer to balance the impact of client delays caused by device heterogeneity or local data heterogeneity. 

Our proposed aggregation strategy is outlined in Algorithm~\ref{algorithm2}, and a comprehensive list of notations used in this paper is listed in Table~\ref{tab:notation}. Upon the receipt of the local model at the PS, we evaluate the obsolescence/staleness of the current model based on the disparity between its and the latest global model's timestamps -- greater disparities signify a "stale" local client model. Moreover, when added to the buffer, stale models are assigned a smaller scaling factor for aggregation. In the case of different local models from the same client being received before reaching the aggregation target in the buffer, a smaller scaling factor will be assigned to the particular high frequency client.
This adjustment aims to prevent performance skewness towards resource-efficient clients, which could otherwise lead to an underrepresented global model.
When the aggregation target is reached, we aggregate the sum of the scaled local models currently stored in the buffer as a new global aggregation model. The scale factor $\beta$ of the $n$-th client within the buffer can be defined as:

\begin{equation}
\label{eq:SF}
\beta_n = \frac{|D_n| \cdot e^{s_n \cdot \frac{1}{f_n}}}{\sum_{b=1}^{B} |D_b| \cdot e^{s_b \cdot \frac{1}{f_b}}}
\end{equation}
where:
\begin{itemize}
    \item \( s = (t - \tau + 1)^{-\alpha} \) represents the model's staleness factor within the buffer layer.
    \item \( f \) represents the model's upload frequency within the buffer layer.
    \item \( B \) is the total number of devices in the buffer layer.
    \item \( |D_n| \) represents the data samples stored on the \( n^{th} \) device.
    \item \(\sum_{b=1}^{B} |D_b| \) is the total number of data samples stored across all \( B \) devices in the buffer layer.
\end{itemize}

\section{Experiments}
\label{sec:experiments}
To evaluate the proposed aggregation strategy (detailed above) within AFL settings, we employ the clothing-based Fashion-MNIST image recognition task for benchmarking~\cite{xiao2017fashion}.
This dataset contains 60,000 training and 10,000 test images, formatted as $28\times28$ grayscale images spanning 10 apparel classes. 
We distribute this training data in both IID and non-IID setups across the local datasets of 10 simulated FL clients and then quantify model performance aggregated from their intermittent updates. 
We compare our dynamic, latency-adjusted aggregation scheme against two state-of-the-art approaches, as introduced in Section~\ref{sec:related}: FedAsync~\cite{xie2019asynchronous} and Papaya~\cite{huba2022papaya}.

In order to assess the most salient bias in AFL (considering the FedAsync method) and its impact on the overall accuracy of the global model. We designed an experimental setup consisting of 10 FL clients (summarised in Table~\ref{table:experiment3}). To mimic hardware disparities, half of the clients were deliberately subjected to resource limitations by introducing delays in their updates. Additionally, we investigated the consequences of data distortion or class imbalance of local client training datasets by varying the number of classes (out of a total of 10) for each client during local training. Finally, an IID experiment, where classes and training data volumes were uniform across all FL clients, was also conducted to illustrate the ``accuracy cost" associated with such modifications. The hyperparameters used for all simulated AFL experiments are consistent and detailed in Table~\ref{tab:hyperparameters}.

\begin{table}[t]
\centering
\caption{Common hyperparameters settings for our simulated AFL experiments.}
\label{tab:hyperparameters}
\begin{tabular}{lc}
\toprule
\textbf{Parameter} & \textbf{Value}  \\
\midrule
Global communication rounds & 100 \\
Scale Factor ($\alpha$) & 0.5 \\
Buffer Aggregation Target & 3 \\
Learning Rate ($\eta$)  & 0.001  \\
Regularization Parameter ($\mu$) & 0.01  \\
Local Training Epoch (E)  & 2  \\
Local Minibatches (b) & 50 \\
Model for Training Task & CNN \\
Optimizer for Training Task & SGD \\
\bottomrule
\end{tabular}
\end{table}

\definecolor{myblue}{RGB}{10, 7, 178} 
\definecolor{myred}{RGB}{199, 10, 10}  
\definecolor{mygreen}{RGB}{10, 150, 10}  

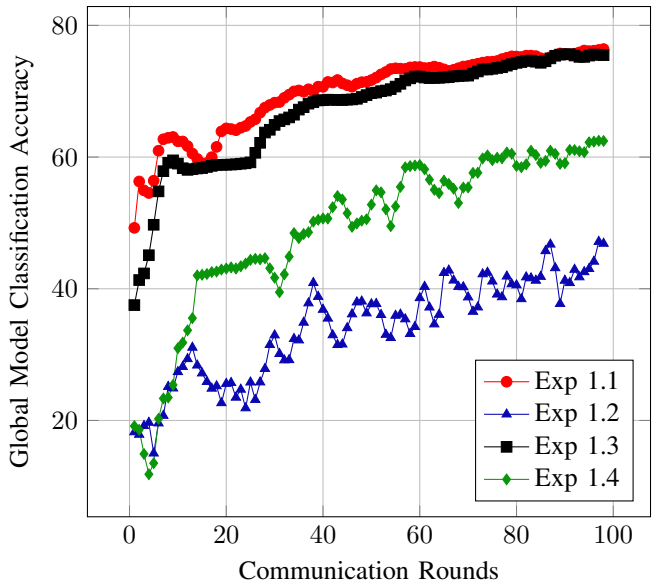
\begin{figure}[t]
\begin{tikzpicture}
\begin{axis}[
    width=0.5\textwidth, 
    height=8.370cm, 
    xlabel={Communication Rounds}, 
    ylabel={Global Model Classification Accuracy}, 
    grid={major},
    legend entries={Exp 1.1, Exp 1.2, Exp 1.3, Exp 1.4},
    legend pos= {south east},
]
\addplot [red, mark=*] table [x=Step, y=iid, col sep=comma] {smoothed_fig4_data.csv};
\addplot [myblue, mark=triangle*] table [x=Step, y=resource_constrained10_noniid, col sep=comma] {smoothed_fig4_data.csv};
\addplot [black, mark=square*] table [x=Step, y=resource_constrained3_noniid, col sep=comma] {smoothed_fig4_data.csv};
\addplot [mygreen, mark=diamond*] table [x=Step, y=fedasync_noniid, col sep=comma] {smoothed_fig4_data.csv};
\end{axis}
\end{tikzpicture}
\label{fig:results_2_smoothed}
\caption{Results plotted for IID versus various Non-IID FL client training dataset distributions settings with their computing resources corresponding to the following units/fractions $[100, 95, 90, 85, 80, 25, 20, 15, 10, 5]$ where the larger values imply better computing resources. Note, that the hyperparameter settings used are shown in Table \ref{tab:hyperparameters}. Exp 1.1: IID client training data, Exp 1.2: Resource constrained devices with 3 training classes, Exp 1.3 Resource constrained devices with all training classes, and Exp 1.4: Client training data with Dirichlet distribution.}
\end{figure}


\begin{table}[t]
\centering
\caption{Experimental hyper-parameter setup for results in Figure \ref{fig:results_1_smoothed}}
\label{table:experiment3}
\begin{tabular}{p{1.1cm}llp{1.5cm}}
\toprule
\textbf{Experiment ID} & \textbf{FL client compute power} & \textbf{data split} & \textbf{Class distribution per client}  \\
\midrule
\textcolor{myred}{1.1} & 50\% resource constrained & IID &  10  \\
\midrule
\multirow{2}{*}{\textcolor{myblue}{1.2}} & 50\% resource constrained & \multirow{2}{*}{non-IID} &  10  \\
 &50\% resource efficient &  &  3  \\
\midrule
\multirow{2}{*}{1.3} & 50\% resource constrained &  \multirow{2}{*}{non-IID} &  3  \\
 &50\% resource efficient & &  10  \\
\midrule
\textcolor{mygreen}{1.4} & 50\% resource constrained & non-IID &  Dirichlet\\
\bottomrule
\end{tabular}
\end{table}

We designate the simulated FL clients as $c_0-c_9$, and their computing capabilities correspond to the following fractions $[100, 95, 90, 85, 80, 25, 20, 15, 10, 5]$, where the larger values implied better computing resources available. The Dirichlet distribution was employed to generate the number of classes available per client to train within the final row (Exp. 1.4).

In our second experimental setup, we evaluated the differences in performance of the global accuracy during AFL training of our proposed method compared with other advanced algorithms such as FedAsync\cite{xie2019asynchronous} and Papaya\cite{huba2022papaya}. We used the same experimental settings as the final row (Exp. num. 1.4) of Table~\ref{table:experiment3} and hyperparameters from Table~\ref{tab:hyperparameters}.


\section{Results}\label{sec:results}

\begin{figure}[t]
\begin{tikzpicture}
\begin{axis}[
    width=0.5\textwidth, 
    height=8.370cm, 
    xlabel={Communication Rounds},
    ylabel={Global Model Classification Accuracy},
    grid=major,
    legend pos=south east,
    legend entries={Our Method, Meta's Papaya, FedASYNC},
]
\addplot [myblue, mark=*] table [x=Step, y=jikun_flasync, col sep=comma] {smoothed_afl_bench_data.csv};
\addplot [myred, mark=square*] table [x=Step, y=meta_papaya, col sep=comma] {smoothed_afl_bench_data.csv};
\addplot [mygreen, mark=triangle*] table [x=Step, y=fedasync, col sep=comma] {smoothed_afl_bench_data.csv};
\end{axis}
\end{tikzpicture}
\label{fig:results_1_smoothed}
\caption{Comparison between \textcolor{myblue}{our proposed method}, \textcolor{myred}{Meta's PAPAYA\cite{huba2022papaya}} and \textcolor{mygreen}{FedAsync\cite{xie2019asynchronous}} AFL systems illustrating classification accuracy using the FashionMNIST dataset under experimental setup 1.4 as described Table \ref{table:experiment3}, and hyperparameter settings shown in Table \ref{tab:hyperparameters}.}
\end{figure}
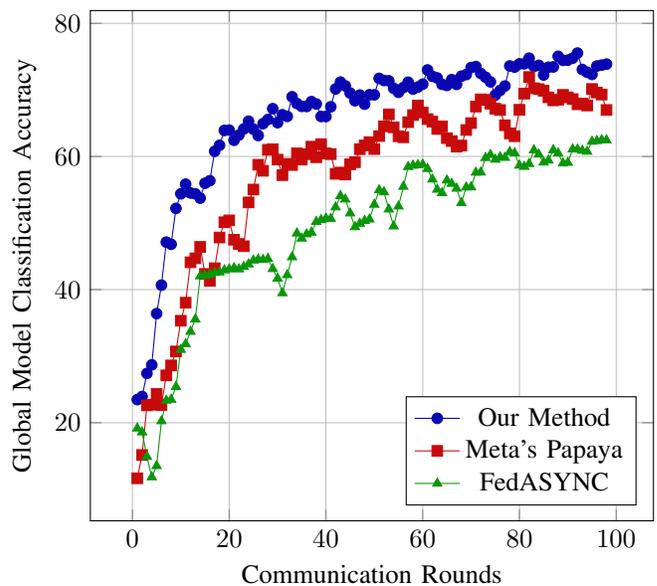




From Fig.~\ref{fig:results_2_smoothed}, it is evident that, in Experiment 1.1, the heterogeneous training devices based on IID data achieve the highest global model accuracy. 
Experiment 1.2 represents computationally resource constrained clients training with access to all 10 classes of the FashionMNIST dataset and the computationally efficient clients training with only 3 classes can achieve almost similar results as the IID setting. 
However, in the reversed scenario (Experiment 1.3), where weaker clients' training is based on all 10 classes and stronger clients train on only 3 classes, this results in the worst performance in terms of classification accuracy.

Experiment 1.4 used the Dirichlet distribution setting, wherein each client was assigned on average 4 or 5 classes of training data (out of a possible 10). This choice was made to emulate a more practical and diverse FL setup, ensuring that no single client had access to the full set of classes to train with. Across these four different experimental settings, it is evident that more powerful clients exert a dominant influence on the training process, and variations in computing resources and training data do introduce bias in the global model. Performance deterioration is particularly noteworthy in scenarios where resource constrained clients have access to the majority classes of the dataset. This can be attributed to the fact that slower clients, with a reduced update frequency, have a lesser contribution to the global model which is heavily biased towards clients with the most amount of updates. 

We next evaluated our proposed method targeting mitigation against such bias and compared its performance with two state-of-the-art AFL methods in PAPAYA and FedAsync.
Global model accuracy for each during AFL training is shown in Fig.~\ref{fig:results_1_smoothed}. We used the simulated experimental setup as described in row $1.4$ of Table \ref{table:experiment3}. The results show that our proposed method improves the FashinMNIST classification performance by more than 10\% and 19\% when compared to the PAPAYA and FedAsync methods respectively (for a fixed set of communication rounds).

\section{Discussion and Conclusion}\label{sec:conclusion}

In this paper, we conducted multiple experiments based on an Asynchronous FL environment to evaluate the biases caused by FL device hardware heterogeneity and non-IID client data distributions.
High-performance devices in these settings quickly train and update models in asynchronous environments, while resource constrained devices may delay submitting their updates. This mismatch can lead to under-consideration of updates for some devices when aggregating global models at the PS, as they are relatively outdated/stale compared to the updated global model state. In addition, network latency and interruptions may also lead to the loss or delay of certain device updates, which poses a threat to the stability and convergence of the global model. 

Moreover, data heterogeneity also negatively impacts performance in such FL applications. The training data stored in the FL client device may vary significantly due to geographical location, user behaviour habits, or other factors. This means that the data of certain devices may not represent the overall data distribution (of all FL clients), or some key subsets of data may only exist on a few devices. Such data distribution may lead to excessive optimisation of the local model on specific tasks or subsets of data, thereby compromising the global ability to generalise to new data or different devices. In terms of data volume, some devices may have rich data (lots of classes, large volume), while others may have relatively few. This imbalance, without proper processing, can cause data-rich devices to have an overwhelming impact on the global model. 

In this work, we show through extensive experimentation that such bias, caused by hardware and data heterogeneity, in AFL, cannot be completely eliminated but can be appropriately reduced. We have proposed a new aggregation method to reduce the impact of such bias on the global model performance. Our method has been experimentally demonstrated to effectively improve the global model accuracy in AFL environments when compared with other state-of-the-art benchmark algorithms. 

For future work, we aim to cover other datasets and different model architectures to evaluate the robustness of our method. Moving forward, promising avenues exist to expand the evaluation of our aggregation technique across even more heterogeneous AFL system deployments. Additional dimensions of hardware diversity could entail sampling various device computation, memory, and power constraints. On the model representation side, assessing performance under conditions of quantisation or pruned neural networks tailored to match the resource limitations of local devices would prove highly valuable. Developing robust, adaptive AFL aggregation schemes capable of harmonising diverse hardware configurations, statistical data divergences, and specialised local model representations remains an open challenge of great practical interest.



\bibliographystyle{IEEEtran}
\bibliography{references}

\begin{thebibliography}{10}
\providecommand{\url}[1]{#1}
\csname url@samestyle\endcsname
\providecommand{\newblock}{\relax}
\providecommand{\bibinfo}[2]{#2}
\providecommand{\BIBentrySTDinterwordspacing}{\spaceskip=0pt\relax}
\providecommand{\BIBentryALTinterwordstretchfactor}{4}
\providecommand{\BIBentryALTinterwordspacing}{\spaceskip=\fontdimen2\font plus
\BIBentryALTinterwordstretchfactor\fontdimen3\font minus \fontdimen4\font\relax}
\providecommand{\BIBforeignlanguage}[2]{{%
\expandafter\ifx\csname l@#1\endcsname\relax
\typeout{** WARNING: IEEEtran.bst: No hyphenation pattern has been}%
\typeout{** loaded for the language `#1'. Using the pattern for}%
\typeout{** the default language instead.}%
\else
\language=\csname l@#1\endcsname
\fi
#2}}
\providecommand{\BIBdecl}{\relax}
\BIBdecl

\bibitem{mcmahan2017communication}
B.~McMahan, E.~Moore, D.~Ramage, S.~Hampson, and B.~A. y~Arcas, ``Communication-efficient learning of deep networks from decentralized data,'' in \emph{Artificial intelligence and statistics}.\hskip 1em plus 0.5em minus 0.4em\relax PMLR, 2017, pp. 1273--1282.

\bibitem{naeem2020sdn}
F.~Naeem, M.~Tariq, and H.~V. Poor, ``Sdn-enabled energy-efficient routing optimization framework for industrial internet of things,'' \emph{IEEE Transactions on Industrial Informatics}, vol.~17, no.~8, pp. 5660--5667, 2020.

\bibitem{ferraguig2021survey}
L.~Ferraguig, Y.~Djebrouni, S.~Bouchenak, and V.~Marangozova, ``Survey of bias mitigation in federated learning,'' in \emph{Conf{\'e}rence francophone d'informatique en Parall{\'e}lisme, Architecture et Syst{\`e}me}, 2021.

\bibitem{konevcny2016federated}
J.~Kone{\v{c}}n{\`y}, H.~B. McMahan, F.~X. Yu, P.~Richt{\'a}rik, A.~T. Suresh, and D.~Bacon, ``Federated learning: Strategies for improving communication efficiency,'' \emph{arXiv preprint arXiv:1610.05492}, 2016.

\bibitem{li2019asynchronous}
Y.~Li, S.~Yang, X.~Ren, and C.~Zhao, ``Asynchronous federated learning with differential privacy for edge intelligence,'' \emph{arXiv preprint arXiv:1912.07902}, 2019.

\bibitem{diao2020heterofl}
E.~Diao, J.~Ding, and V.~Tarokh, ``Heterofl: Computation and communication efficient federated learning for heterogeneous clients,'' \emph{arXiv preprint arXiv:2010.01264}, 2020.

\bibitem{xu2021asynchronous}
C.~Xu, Y.~Qu, Y.~Xiang, and L.~Gao, ``Asynchronous federated learning on heterogeneous devices: A survey,'' \emph{arXiv preprint arXiv:2109.04269}, 2021.

\bibitem{li2020federated}
T.~Li, A.~K. Sahu, M.~Zaheer, M.~Sanjabi, A.~Talwalkar, and V.~Smith, ``Federated optimization in heterogeneous networks,'' \emph{Proceedings of Machine learning and systems}, vol.~2, pp. 429--450, 2020.

\bibitem{li2014communication}
M.~Li, D.~G. Andersen, A.~J. Smola, and K.~Yu, ``Communication efficient distributed machine learning with the parameter server,'' \emph{Advances in Neural Information Processing Systems}, vol.~27, 2014.

\bibitem{zhao2018federated}
Y.~Zhao, M.~Li, L.~Lai, N.~Suda, D.~Civin, and V.~Chandra, ``Federated learning with non-iid data,'' \emph{arXiv preprint arXiv:1806.00582}, 2018.

\bibitem{beutel2020flower}
D.~J. Beutel, T.~Topal, A.~Mathur, X.~Qiu, J.~Fernandez-Marques, Y.~Gao, L.~Sani, K.~H. Li, T.~Parcollet, P.~P.~B. de~Gusm{\~a}o \emph{et~al.}, ``Flower: A friendly federated learning research framework,'' \emph{arXiv preprint arXiv:2007.14390}, 2020.

\bibitem{xie2019asynchronous}
C.~Xie, S.~Koyejo, and I.~Gupta, ``Asynchronous federated optimization,'' \emph{arXiv preprint arXiv:1903.03934}, 2019.

\bibitem{chen2021fedsa}
M.~Chen, B.~Mao, and T.~Ma, ``Fedsa: A staleness-aware asynchronous federated learning algorithm with non-iid data,'' \emph{Future Generation Computer Systems}, vol. 120, pp. 1--12, 2021.

\bibitem{chen2021towards}
Z.~Chen, W.~Liao, K.~Hua, C.~Lu, and W.~Yu, ``Towards asynchronous federated learning for heterogeneous edge-powered internet of things,'' \emph{Digital Communications and Networks}, vol.~7, no.~3, pp. 317--326, 2021.

\bibitem{chen2019efficient}
M.~Chen, B.~Mao, and T.~Ma, ``Efficient and robust asynchronous federated learning with stragglers,'' in \emph{International Conference on Learning Representations}, 2019.

\bibitem{huba2022papaya}
D.~Huba, J.~Nguyen, K.~Malik, R.~Zhu, M.~Rabbat, A.~Yousefpour, C.-J. Wu, H.~Zhan, P.~Ustinov, H.~Srinivas \emph{et~al.}, ``Papaya: Practical, private, and scalable federated learning,'' \emph{Proceedings of Machine Learning and Systems}, vol.~4, pp. 814--832, 2022.

\bibitem{nguyen2022federated}
J.~Nguyen, K.~Malik, H.~Zhan, A.~Yousefpour, M.~Rabbat, M.~Malek, and D.~Huba, ``Federated learning with buffered asynchronous aggregation,'' in \emph{International Conference on Artificial Intelligence and Statistics}.\hskip 1em plus 0.5em minus 0.4em\relax PMLR, 2022, pp. 3581--3607.

\bibitem{xiao2017fashion}
H.~Xiao, K.~Rasul, and R.~Vollgraf, ``Fashion-mnist: a novel image dataset for benchmarking machine learning algorithms,'' \emph{arXiv preprint arXiv:1708.07747}, 2017.

\end{thebibliography}
\clearpage

\end{document}